\documentclass[sigconf]{acmart}
\AtBeginDocument{%
  \providecommand\BibTeX{{%
    \normalfont B\kern-0.5em{\scshape i\kern-0.25em b}\kern-0.8em\TeX}}}

\copyrightyear{2024}
\acmYear{2024}
\setcopyright{rightsretained}
\acmConference[CHI EA '24]{Extended Abstracts of the CHI Conference on
Human Factors in Computing Systems}{May 11--16, 2024}{Honolulu, HI, USA}
\acmBooktitle{Extended Abstracts of the CHI Conference on Human Factors in
Computing Systems (CHI EA '24), May 11--16, 2024, Honolulu, HI, USA}
\acmDOI{10.1145/3613905.3650788}
\acmISBN{979-8-4007-0331-7/24/05}

\usepackage{algorithm}
\usepackage{algpseudocode}
\usepackage{algorithmicx}
\usepackage{multirow}
\usepackage{array}
\usepackage{subcaption}

\newcolumntype{P}[1]{>{\centering\arraybackslash}p{#1}}

\acmSubmissionID{123-A56-BU3}

\begin{document}

\makeatletter
\def\@ACM@checkaffil{
    \if@ACM@instpresent\else
    \ClassWarningNoLine{\@classname}{No institution present for an affiliation}%
    \fi
    \if@ACM@citypresent\else
    \ClassWarningNoLine{\@classname}{No city present for an affiliation}%
    \fi
    \if@ACM@countrypresent\else
        \ClassWarningNoLine{\@classname}{No country present for an affiliation}%
    \fi
}
\makeatother




\title{iCONTRA: Toward Thematic Collection Design Via Interactive Concept Transfer}


\author{Dinh-Khoi Vo}
\authornote{Equal contribution}
\orcid{0000-0001-8831-8846}
\affiliation{%
  \institution{\textsuperscript{\rm 1} University of Science, VNU-HCM, Ho Chi Minh City, Vietnam}
}
\affiliation{
  \institution{\textsuperscript{\rm 2} Vietnam National University, Ho Chi Minh City, Vietnam}
}

\author{Duy-Nam Ly}
\authornotemark[1]
 \orcid{0000-0003-4304-2334}
\affiliation{%
  \institution{\textsuperscript{\rm 1} University of Science, VNU-HCM, Ho Chi Minh City, Vietnam}
}
\affiliation{
  \institution{\textsuperscript{\rm 2} Vietnam National University, Ho Chi Minh City, Vietnam}
}

\author{Khanh-Duy Le}
\orcid{0000-0002-8297-5666}
\affiliation{%
  \institution{\textsuperscript{\rm 1} University of Science, VNU-HCM, Ho Chi Minh City, Vietnam}
}
\affiliation{
  \institution{\textsuperscript{\rm 2} Vietnam National University, Ho Chi Minh City, Vietnam}
}

\author{Tam V. Nguyen}
\orcid{0000-0003-0236-7992}
\affiliation{%
  \institution{\textsuperscript{\rm 3} Department of Computer Science University of Dayton \\Ohio, United States}
}

\author{Minh-Triet Tran}
\orcid{0000-0003-3046-3041}
\affiliation{%
  \institution{\textsuperscript{\rm 1} University of Science, VNU-HCM, Ho Chi Minh City, Vietnam}
}
\affiliation{
  \institution{\textsuperscript{\rm 2} Vietnam National University, Ho Chi Minh City, Vietnam}
}

\author{Trung-Nghia Le}
\authornote{Corresponding author. Email address: ltnghia@fit.hcmus.edu.vn}
\orcid{0000-0002-7363-2610}
\affiliation{%
  \institution{\textsuperscript{\rm 1} University of Science, VNU-HCM, Ho Chi Minh City, Vietnam}
}
\affiliation{
  \institution{\textsuperscript{\rm 2} Vietnam National University, Ho Chi Minh City, Vietnam}
}


\begin{abstract}

Creating thematic collections in industries demands innovative designs and cohesive concepts. Designers may face challenges in maintaining thematic consistency when drawing inspiration from existing objects, landscapes, or artifacts. While AI-powered graphic design tools offer help, they often fail to generate cohesive sets based on specific thematic concepts. In response, we introduce iCONTRA, an interactive CONcept TRAnsfer system. With a user-friendly interface, iCONTRA enables both experienced designers and novices to effortlessly explore creative design concepts and efficiently generate thematic collections. We also propose a zero-shot image editing algorithm, eliminating the need for fine-tuning models, which gradually integrates information from initial objects, ensuring consistency in the generation process without influencing the background. A pilot study suggests iCONTRA's potential to reduce designers' efforts. Experimental results demonstrate its effectiveness in producing consistent and high-quality object concept transfers. iCONTRA stands as a promising tool for innovation and creative exploration in thematic collection design. The source code will be available at:~\url{https://github.com/vdkhoi20/iCONTRA}.

\end{abstract}

\ccsdesc[500]{Human-centered computing~Human computer interaction (HCI)}
\ccsdesc[500]{Human-centered computing~Interactive systems and tools}
\ccsdesc[500]{Computing methodologies~Machine learning}

\keywords{Thematic collection design, Zero-shot image editing, Diffusion model}

\begin{teaserfigure}
    \includegraphics[width=\textwidth]{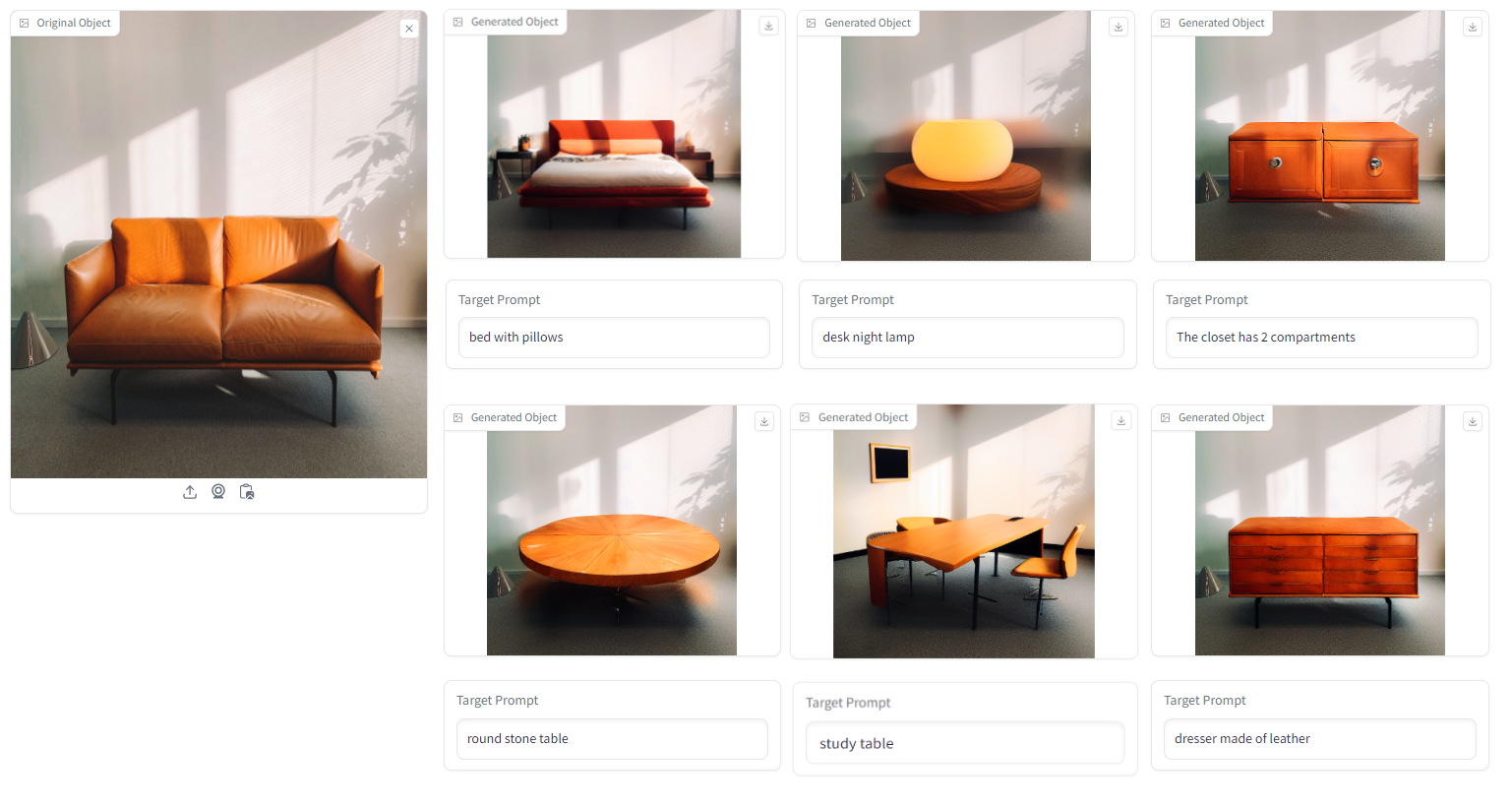}
  \caption{Given a single object image (i.e., left image), when users input brief prompts, our proposed iCONTRA system unleashes the power of generative AI, recommending objects that have the same theme as the input object (i.e., right images). The system also preserves the original structure and scene layout with unwavering precision, which is helpful for designers when visualizing new objects in the same environment, such as background and light.}
  \label{fig:teaser}
\end{teaserfigure}

\maketitle

\section{Introduction}

Crafting objects with a unified concept or theme is a critical aspect of various industries, heavily dependent on the creativity of designers for innovative designs, particularly to meet the demand for cohesive concepts in areas such as fashion and interior decor. Designers often face challenges where they must draw inspiration from existing objects, landscapes, or artifacts to achieve thematic cohesion in their design collections. For example, a designer may aspire to fashion a clothing line influenced by the intricate patterns found in a vintage piece of furniture or the harmonious color palette of a serene natural landscape. In such cases, the ability to seamlessly transfer these inspirations into cohesive and stylish fashion designs becomes crucial. This intricate process underscores the designer's proficiency in conceptualization and innovation.

More specific, when crafting thematic objects, designers can explore different websites, such as Pinterest, Instagram, using keyword descriptions to find inspiration and reference for creating a collection with a unified design theme. For instance, given a sofa placing in a glass window showroom, designers need to create other furniture, including bed, table, wardrobe that share the same design concept as the sofa, as illustrated in Fig.~\ref{fig:teaser}. However, this process can be time-consuming and challenging, often resulting in difficulty finding suitable objects while potentially losing the feeling of layout and environment of the original image. 

AI-driven graphic design tools redefine the creative landscape by leveraging AI to streamline and enhance the design process~\cite{le2023vides}. These tools not only expedite tasks but also serve as innovative aids, saving time and providing inspiration. With capabilities ranging from swift task execution to offering creative suggestions, AI becomes a valuable partner for designers, unlocking unlimited possibilities and pushing creative boundaries beyond traditional approaches. While existing text-driven image manipulation methods excel in tasks like translation~\cite{liu2017unsupervised,Ruiz_2023_CVPR,huang2018multimodal} and style transfer~\cite{li2017demystifying,8732370}, challenges arise due to the lack of specific object shape targets, making the process time-consuming and requiring fine-tuning efforts. On the other hand, AI-powered commercial softwares such as Adobe Firefly
~\footnote{\url{https://www.adobe.com/products/firefly.html}}
, Midjourney
~\footnote{\url{https://www.midjourney.com/}}
, DALL$\cdot$E 2
~\footnote{\url{https://openai.com/dall-e-2}}
, DALL$\cdot$E 3
~\footnote{\url{https://openai.com/dall-e-3}}
, Stable Diffusion 
~\footnote{\url{https://platform.stability.ai/sandbox/text-to-image}} 
are not freely available for everyone and may be difficult to use for people without AI knowledge. Moreover, these tools cannot directly support designing thematic collections. These tools tend to produce entirely different objects, resulting in a loss of the original environmental background and a deviation from the desired thematic continuity.

To overcome these challenges, we introduce a novel interactive CONcept TRAnsfer (iCONTRA) system, designed to assist both experienced designers and individuals without prior design skills. iCONTRA aims to provide a user-friendly platform for creative exploration and expression in the design process of thematic collection. Our system can address limitations of existing systems in creating cohesive sets of objects. The user interface provides a seamless experience, allowing users to upload the original image and provide textual descriptions in each generation cell for the desired object (See Fig.~\ref{fig:InterfaceGuild}). With six cells available, users can generate multiple objects iteratively. After generating initial objects, users can further refine them using convenient import and edit prompt features. They can also observe the generated object alongside the edited prompt, facilitating the achievement of desired designs. 
This iterative and user-friendly approach enhances the overall design exploration process, providing creative freedom and ease of use while reducing the effort required to find similar patterns, thus enhancing the overall user experience.



Our proposed iCONTRA system is built on a cutting-edge generative model~\cite{rombach2021highresolution}, indicating its capacity to generate diverse and high-quality images. This model is tailored to achieve consistent and intricate non-rigid object generation while preserving overall textures and identity from the original object. When replacing an object in an image by another one, the central challenge is preserving the original information of the real object, in which previous methods encountered difficulties~\cite{hertz2022prompt,parmar2023zeroshot,chefer2023attendandexcite, elarabawy2022direct}. To offer more robust solution, we develop a novel zero-shot image editing model allowing replacing objects in an image without the need for fine-tuning or specific training. Our algorithm exploits information from the original image, preventing abrupt changes during the generation process. Additionally, we automatically mask the foreground object from the background, minimizing the impact of background adjustment. This enables querying correlated local structures and textures from the original object, ensuring consistency in generating new objects. 


We conducted a pilot study to elicit preliminary insights and feedback on the current state of our system. Through the study sessions, we garnered positive feedback regarding the efficacy of iCONTRA in assisting users to create related objects with the same concept. Additionally, participants highlighted numerous valuable possibilities for the incoming improvement version. The source code will be available at:~\url{https://github.com/vdkhoi20/iCONTRA}.

Our contributions can be summarized as follows:

\begin{itemize}
    \item We propose interactive CONcept TRAnsfer (iCONTRA), an intuitive application designed to effortlessly generate a sequence of cohesive objects with shared conceptual bases, offering a versatile and efficient tool for creative design exploration.
    \item We develop a novel zero-shot image editing algorithm that eliminates the need for training, allowing for the consistent and intricate generation of non-rigid object images. Our approach can preserve the object's characteristics and texture seamlessly without influencing the background.
    \item Insights from a pilot study with participants proficient in design suggest that our system reduces designers' effort in generating desired objects. Experimental results demonstrate the effectiveness of iCONTRA in transferring consistent and high-quality object concepts, showcasing its potential for innovative design applications.
\end{itemize}

\section{Related Work}
\subsection{Text-to-Image}


The realm of text-to-image generation witnessed significant progress due to diffusion-based techniques, such as GLIDE~\cite{nichol2022glide}, DALL$\cdot$E 2~\cite{ramesh2022hierarchical}, and Imagen~\cite{saharia2022photorealistic}. These models employed text embeddings from large language models, exhibiting the ability to generate diverse, high-quality images that aligned with intricate textual prompts. GLIDE and DALL$\cdot$E 2 are conditioned on CLIP textual embeddings while DALL$\cdot$E 2 generates image embeddings from the input text CLIP embedding, followed by image generation through another diffusion model. To handle high-resolution image generation, both GLIDE and Imagen generated low-resolution text-conditioned images using cascaded diffusion models. Rombach~et al.~\cite{rombach2022highresolution} proposed conducting conditional text-to-image diffusion in a reduced-dimensional latent space for efficient training and sampling. Building upon LDM, Stable Diffusion~\cite{rombach2021highresolution} introduced a substantial text-to-image model, trained on a vast dataset, and made available for open research.


Recent computer graphic tools utilizing advanced generative models for design generation and editing have gained significant attention due to their impressive features and user-friendly interfaces. Midjourney 
represents an advanced AI image generator that opened up artistic possibilities. By accepting text prompts and utilizing a Discord bot, it facilitates the creation of detailed and high-quality graphics suitable for both personal and professional projects. DALL$\cdot$E 3
, developed by OpenAI, stands out as a sophisticated image generation model. This tool assists users in creating a diverse range of visuals, from realistic images to stylized illustrations, translating textual descriptions into captivating artwork. Jasper Art
~\footnote{\url{https://www.jasper.ai/art}}
, a creative tool that turned ideas into visual representations,  leverages cutting-edge AI technology to produce unique and one-of-a-kind art based on written input. Adobe Firefly 
is at the forefront of generative AI tools, aiming to revolutionize how creators, designers, and artists engaged with digital content creation. This innovative platform seamlessly transforms textual descriptions into vibrant images, turned sketches into fully realized pictures, and interpreted 3D models into stunning visuals. While existing tools generally performs well in typical cases, in this particular task, users often need to provide detailed descriptions or require visual guidance to generate the desired object, a scenario where these tools commonly face challenges.

\subsection{Text-based Image Editing}
While DiffusionCLIP~\cite{Kim_2022_CVPR} introduced step-by-step diffusion inversion for text-guided image editing, relying on diffusion model refinement, Prompt-to-Prompt~\cite{hertz2022prompt} achieved comprehensive text-guided image editing without diffusion model refinement. It incorporated both global and local editing without predefined masks, primarily focusing on generated image editing due to the unreliability of step-by-step inversion for real images, especially with larger classifier-free guidance scales. Some methods utilized cross-attention or spatial features for editing but often preserved the original layout and struggled with non-rigid transformations. Null-text inversion (NTI)~\cite{mokady2022null} proposed optimal image-specific null-text embeddings for accurate reconstruction, combined with PTP techniques for real image editing. Imagic~\cite{kawar2023imagic}, a related work, facilitated various non-rigid image editing by altering prompts directly, demanding meticulous optimization of textual embeddings and model fine-tuning, making it less user-friendly for ordinary users.

In contrast to Masactrl~\cite{cao_2023_masactrl}, a tuning-free method achieved complex non-rigid and consistent text-guided image editing by replacing self-attention with mutual self-attention, allowing it to query correlated local structures and textures from a source noise for consistency. Building upon the Masactrl approach, our innovative method incorporates a FiOII attention mechanism, enabling consistent and detailed non-rigid image synthesis and editing. Unlike Masactrl, our approach effectively addresses the issue of information loss caused by DIM inversion in real images. Moreover, it can alter a wide range of object attributes, such as pose, shape, and color, by simply modifying the text prompt. Remarkably, these modifications occur without any changes to the model configuration or system architecture, eliminating the need for fine-tuning.

\section{Proposed System}

\subsection{User Interface}

\begin{figure}[!t]
    \centering
    \includegraphics[width=\linewidth]{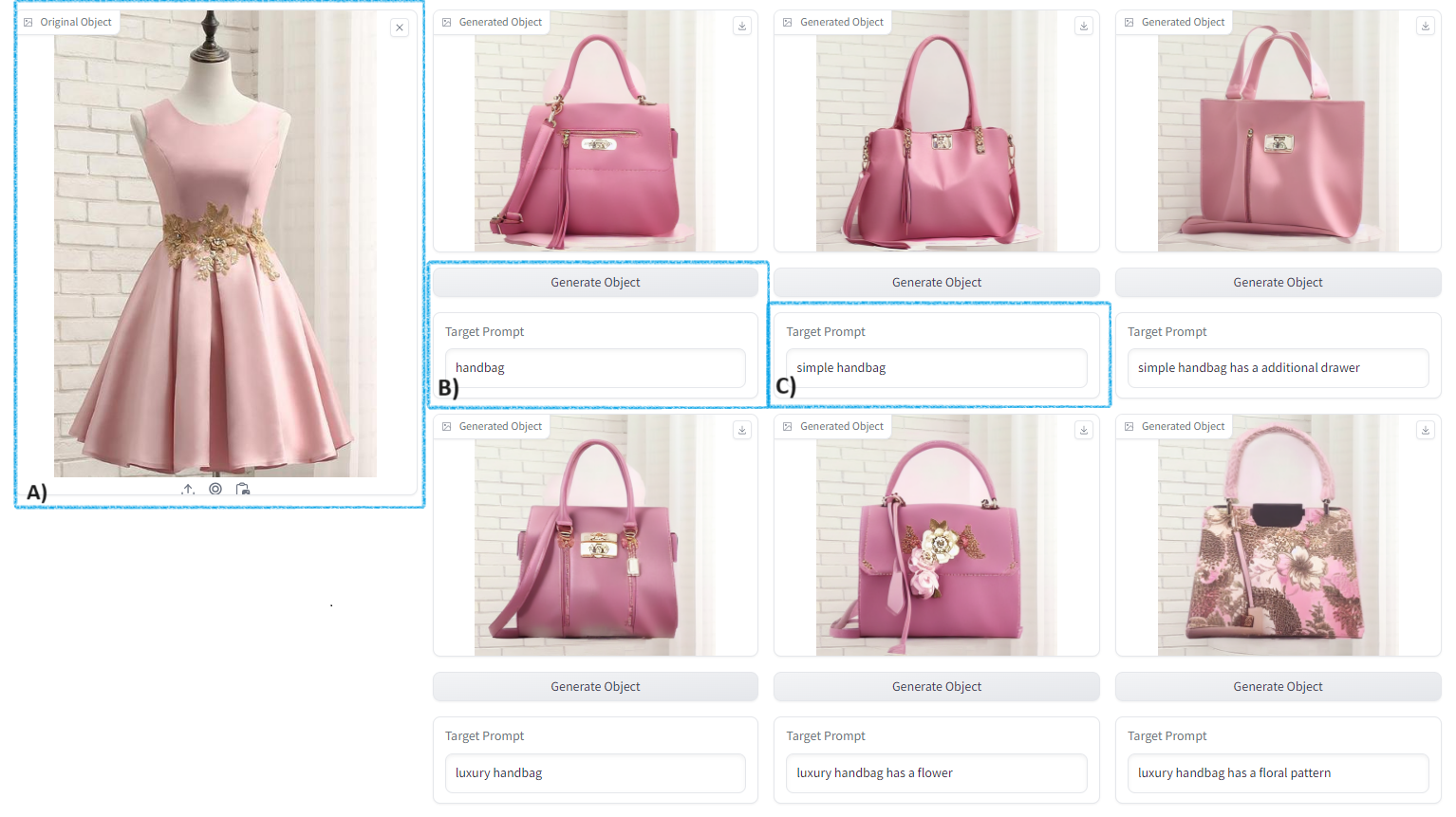}
    \caption{iCONTRA interface. a) Upload the original object image, b) Generate the Object via prompt, c) Modify the prompt until obtain the desired object.} 
    \label{fig:InterfaceGuild}
\end{figure}

In the development of the iCONTRA prototype, we explore several designs, including a conversation chat design like Midjourney or BingAI. Nevertheless, this design is ill-suited for our target users, who seek to create numerous designs from an original image. When entering various prompts to generate multiple designs, users may face the challenge that the back-end responses from the aforementioned tools no longer rely on information from the image users initially input but rather on the context of recent prompts. For instance, an individual employs BingAI to locate a handbag that complements her dress. She uploads the original photo of the dress and inputs numerous prompts to generate multiple designs for consideration. However, after a certain number of prompts, BingAI's responses may stray from the original image information she initially uploaded and instead be influenced by recent prompts. Consequently, she needs to repeatedly re-upload the original image to complete her task.

Therefore, we propose a web-based user interface which offers a seamless experience, ensuring simplicity and efficiency. Users can effortlessly navigate the platform by uploading the original object image on the left-hand side and providing a textual description in each cell on the right-hand side for the desired object (refer to Fig.~\ref{fig:InterfaceGuild}). The interface comprises six cells on the right-hand side, allowing users to generate multiple objects iteratively. The generated objects are guaranteed to be based on the original image and information from the prompts that the user has entered. This leads to users avoiding the necessity of repeatedly re-uploading the original image multiple times, unlike the process with other tools implementing a conversational chat interface. After users generate their initial objects, they can choose to retain and refine them further using the convenient import and edit prompt feature. Additionally, they can observe the generated object alongside the edited prompt to achieve their desired object more effectively. For instance, if a user generates a bag and wishes to add patterns or modify its size, they can seamlessly continue the design process by importing the initial bag and refining it with additional prompts. The intuitive design and straightforward process make it accessible for both experienced designers and those with a limited design background, enhancing the overall user experience. This simplicity reduces the effort needed to find similar patterns on the internet or through other tools.

To expedite the development of a web interface for the initial iCONTRA prototype, we employed Gradio as the foundational technology to construct an interactive interface for users. This interface operates under the logic of the backend, powered by our proposed zero-shot image editing algorithm.

\begin{figure}[!t]
    \centering
    \includegraphics[width=\linewidth]{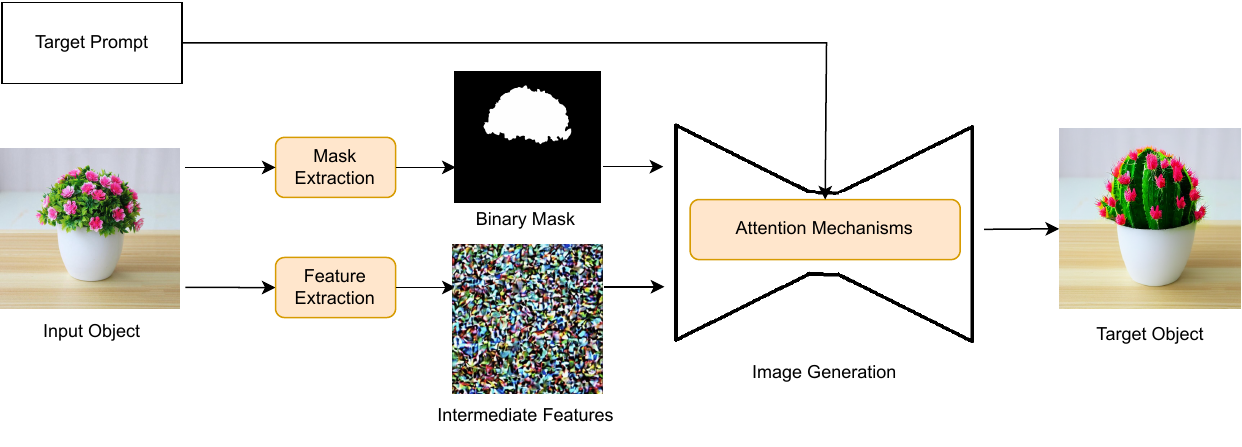}
    \caption{The proposed zero-shot image editing pipeline consists of two phases: data extraction and image synthesis.}
    \label{fig:pipeline}
\end{figure}

\subsection{Zero-Shot Image Editing Algorithm}

The primary objective is to create a target object that conforms to the edited text prompt, maintaining the object contents from while keeping the background environment intact, all without the need for training models. Our proposed zero-shot image editing aims to accomplish spatially edited image synthesis using an input object image, a target prompt, and an automatically derived object mask, as depicted in Fig.~\ref{fig:pipeline}. We build our algorithm on the Stable Diffusion~\cite{rombach2021highresolution} model. The input image in this task often contains only one object or a salient object, so we decide to obtain the object mask using the Rembg library in Python or by allowing the user to type additional prompts corresponding to the desired object and then automatically obtaining the object mask using a LangSam model~\cite{kirillov2023segment}. Similar to Stable Diffusion~\cite{rombach2021highresolution}, we also generate an attention mask based on the target prompt, namely target mask. In the generation phase, the algorithm controls the fade-in of original object's content by gradually interpolating intermediate features guided by the object mask and the target mask via attention mechanisms. As the result, we can edit the object without effecting to background.

\begin{figure*}[!t]
  \centering
  \includegraphics[width=\textwidth]{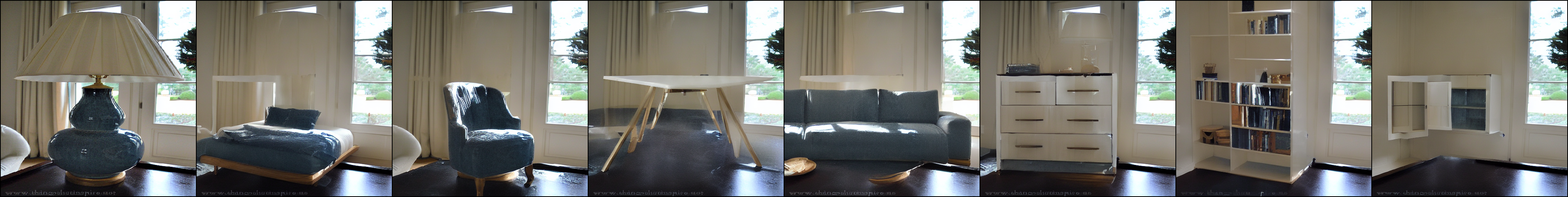}
  \includegraphics[width=\textwidth]{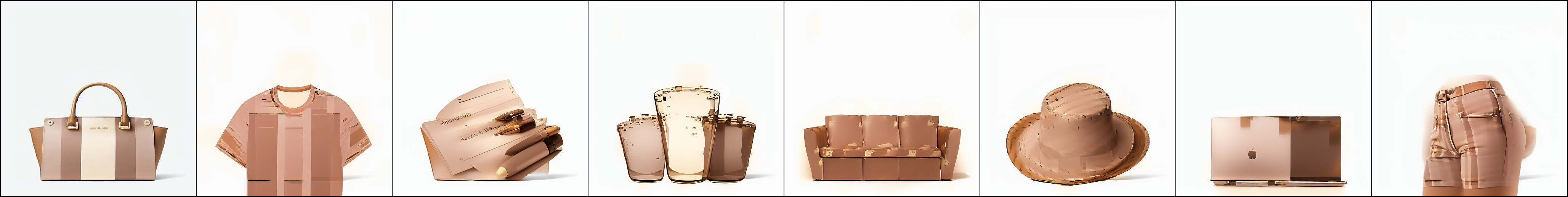}
  \includegraphics[width=\textwidth]{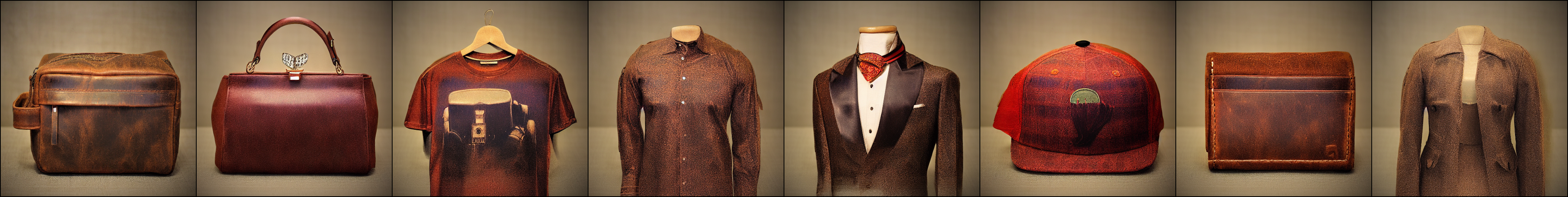}
    \vspace{-5mm}

  \includegraphics[width=\textwidth]{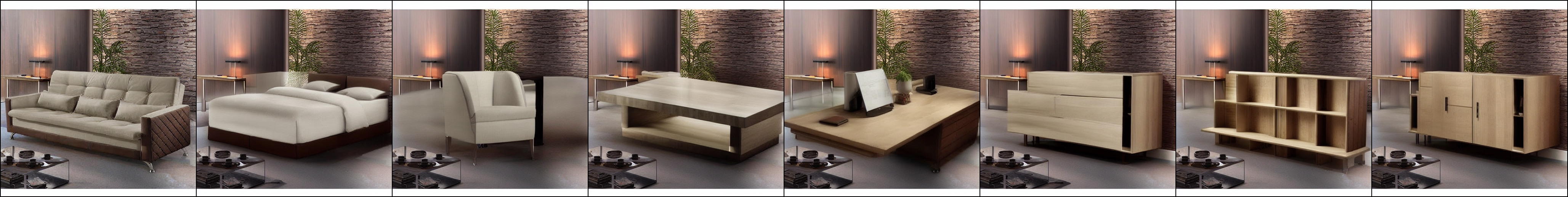}

  \caption{The first column represents the original objects (lamp, bag, wallet and sofa). The series of objects generated share the same concept, starting from the second column.}
  \Description{visualize the results.}
  \label{fig:figvl}
\end{figure*}

We present qualitative images to showcase the performance of iCONTRA. As depicted in Fig.~\ref{fig:figvl}, iCONTRA demonstrates its ability to generate multiple objects with the same concept while preserving the background environments effectively. The results underscore the system's capacity to produce diverse and coherent designs, indicating its potential for creative applications in design spaces. iCONTRA can produce different shapes, colors, etc., consistent with the original object. As seen in the first row with the lamp, it seamlessly integrates the lampshade into the background, ensuring consistent, non-rigid generation across various categories and shapes.



\subsection{Implementation}
We apply our proposed method to the cutting-edge text-to-image Stable Diffusion model, utilizing publicly available version 1.5 checkpoints. Initially, we transform the object image into its base noise map using the deterministic inversion technique of DDIM with null-text guided because problem of image reconstruction problem~\cite{mokady2022null}. During the sampling process, we employ DDIM sampling with 50 denoising iterations, and the classifier-free guidance is set at 7.5 and control with new mechanism fade-in of original object's content by gradually interpolating intermediate features guided by masks.



\begin{figure*}[!t]
  \centering
  
    \includegraphics[width=\textwidth]{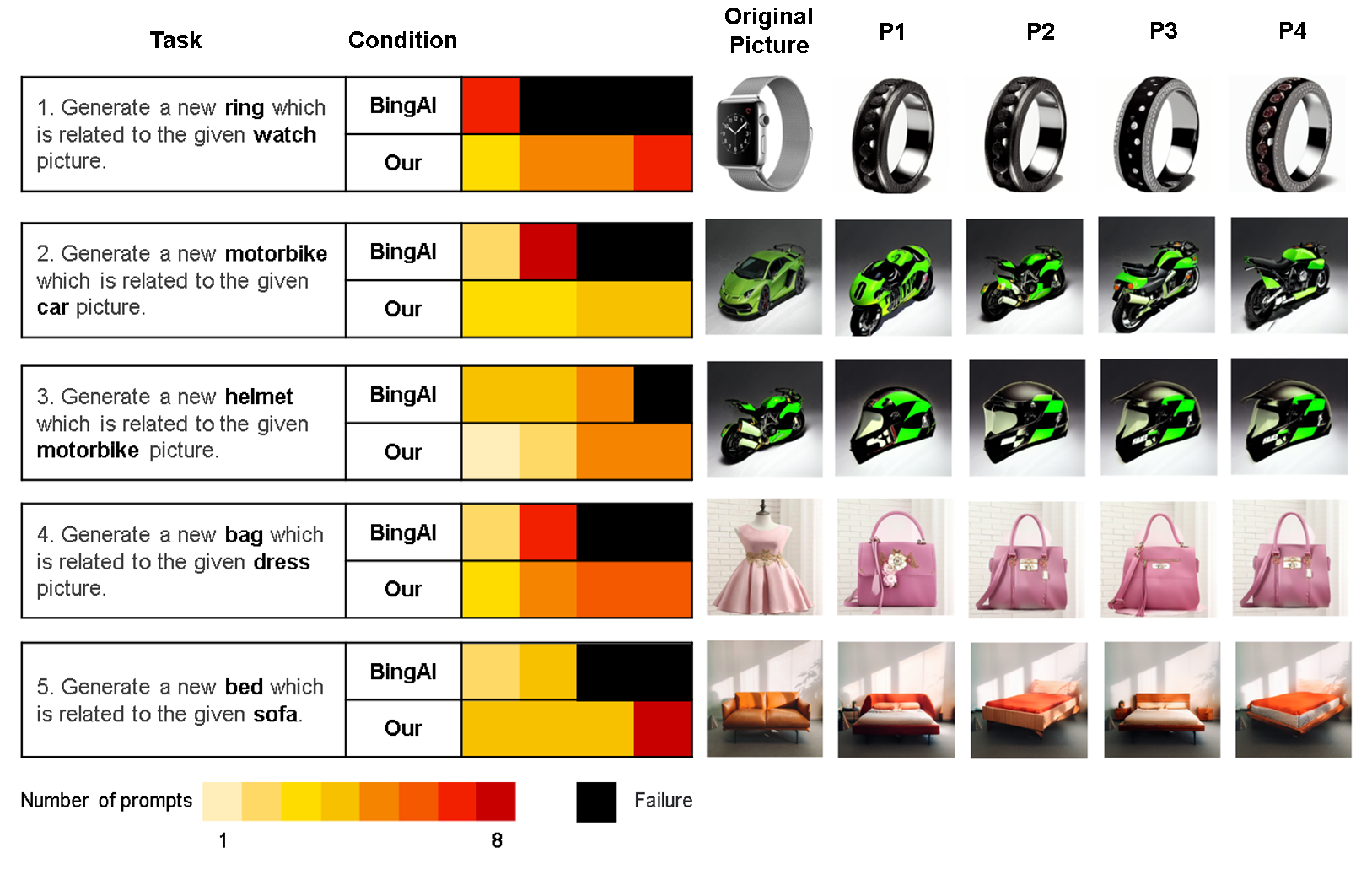}
 
  \vspace{-7mm}
  \caption{Quantitative data of the pilot study. Satisfying results participants generated using iCONTRA.}
  \label{fig:study_results}
\end{figure*}

\section{Pilot Study}
We conducted a preliminary user study to explore how our system might assist users to create families of related objects, especially in comparisons with existing AI-based generation tools. This should provide us early insights on the merits and drawbacks of the system and how we can improve it in the next steps.

\subsection{Participants}
We invited 4 students (2 males, 2 females, average age: 19) from our local university's design club to participate in our study. They are all fresh designers and regularly join in the design activities organized by their club. Three of them are knowledgeable and have used AI tools several times, while the remaining man frequently uses the Midjourney tool to get ideas for his work. 
\subsection{Baseline condition and Measurement}
During our research, we identified BingAI, Midjourney, and Photoshop as relevant tools with functionalities that assist users in generating images based on prompts, which are related to our work. Among these, BingAI stands out as a widely used and free tool, well-known to our participants. Notably, BingAI enables users to upload photos for queries. Consequently, we select BingAI as the baseline for comparison with our system. To assess the effectiveness of iCONTRA in aiding users to create a new object related to an existing one, we recorded the number of prompts users entered until achieving the desired result under both conditions.
\subsection{Tasks}
Before starting the pilot study, we group-interviewed four participants to discuss the typical tasks they are required to undertake. P1 highlighted that in the realm of car racing, it is customary to witness athletes embracing a consistent thematic approach. Additionally, he noted that individuals often acquire accessories that harmonize with the color of their cars. Meanwhile, P2 mentioned that for girls, encountering the challenge of finding a handbag that complements their dress is a common issue. To meet the number of tasks we need, we proposed exploring two topics: jewelry and interior. The consensus among everyone was in favor of these topics. Subsequently, we have designed five tasks as outlined below:
\begin{itemize}
  \item Task 1: Generate a new ring which is related to the given watch picture.
  \item Task 2: Generate a new motorbike which is related to the given car picture.
  \item Task 3: Generate a new helmet which is related to the given motorbike picture.
  \item Task 4: Generate a new bag which is related to the given dress picture.
  \item Task 5: Generate a new bed which is related to the given sofa.
\end{itemize}

\subsection{Apparatus and Procedure}
Our pilot study took place in our lab, the participants performed the tasks we gave using the laptop we provided under our observation. To minize the learning effect in our study, we required two users to try our system first, while the remaining two started with BingAI for their study sessions. At the end of each session, we also gathered feedback and suggestions for the future version of iCONTRA. The overall time for these study sessions was around 40 minutes (including post-interview). The whole pilot study was video-recorded for our data analysis.

\subsection{Quantitative result and Qualitative feedback}



The table above shows that it is difficult for participants to find designs that meet their expectations using BingAI. To explain this, users clearly outlined it in the post-interview. Participants stated that the results response from BingAI are mostly unrealistic. To elicit desired outcomes, participants often needed to provide detailed descriptions of various attributes such as colour and material for the input image. Notably, only in three cases did users achieve the desired results after making two queries with BingAI. Based on our observations, when users repeatedly input prompts and received results that deviate from the original image, they often resorted to re-uploading the initial image. Alternatively, if the expected results were not achieved after a considerable number of prompt entries (at least 8 prompts), users stopped and expressed that they failed in their attempts.

\begin{figure*}[!t]
  \centering
  
  \includegraphics[width=\textwidth]{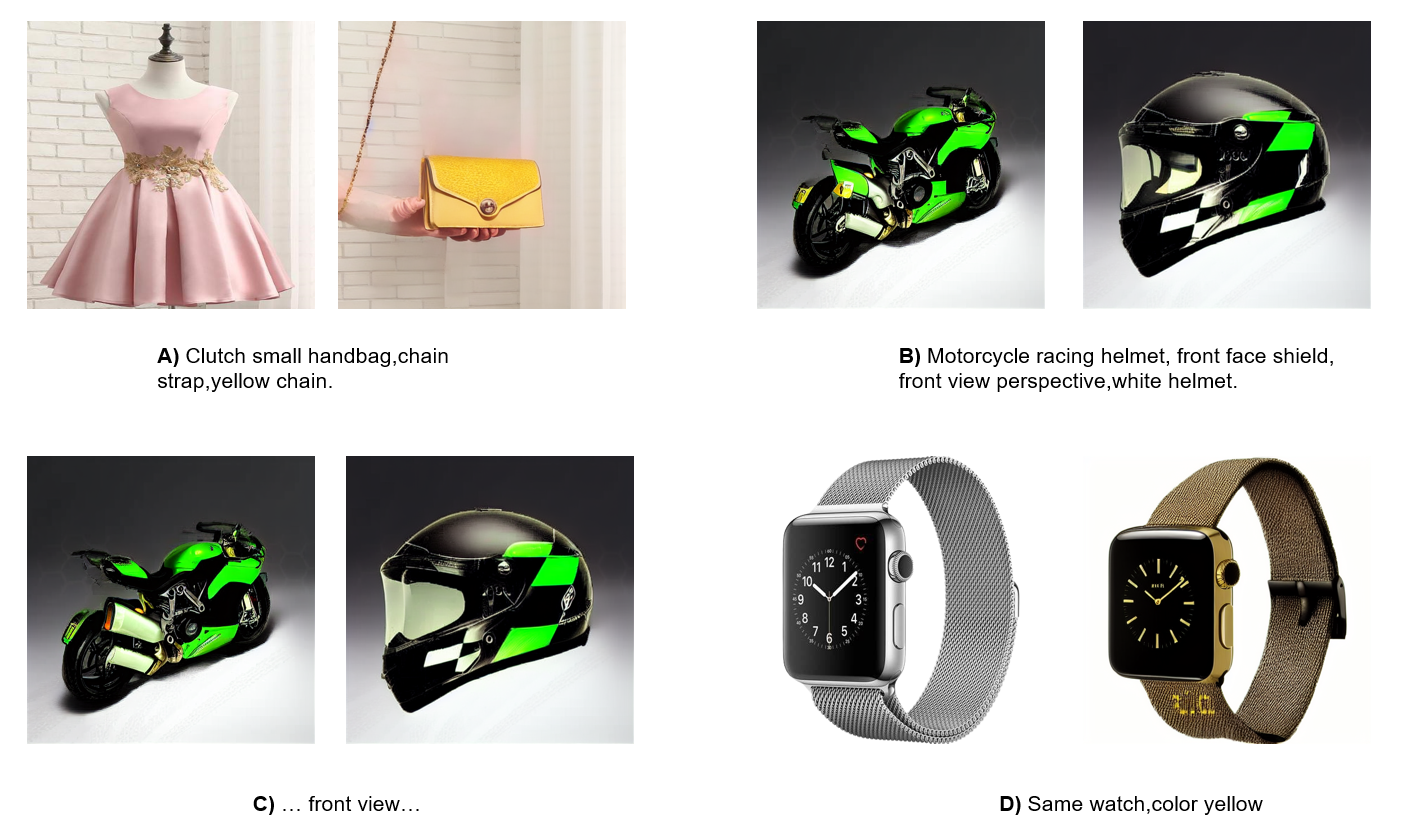}
  \vspace{-9mm}
  \caption{Failure cases which users face. A) The user desires a handbag with a yellow chain, but the model inadvertently changes the entire handbag to yellow. B) The model currently struggles to fully understand the prompt, particularly when it includes a description of a white color helmet at the end of the input text. C) The model is unable to generate different of pose or view of object. D) The model is unable to retain the original object. }
  \label{fig:failure_cases}
\end{figure*}

Meanwhile, all users verbally reported a satisfaction level exceeding 7 out of 10 with the final output image generated using iCONTRA. Through the utilization of iCONTRA, users found themselves requiring less effort to contemplate and describe extensive information about the original object, in contrast to their experience with the baseline tool. However, there were still many cases where users needed to enter prompts at least five times, surpassing our initial expectations. Reviewing the recorded video we found that there were many prompts for participants who wanted to change the perspective or transform a specific part of the object in the resulting image. That led users to spend more prompts on iCONTRA to complete the tasks. We discussed with participants how to tackle these problems and get more insights for the next improvements of iCONTRA. For the need to modify a specific area in the picture, participants expressed a preference for the ability to select the region for alteration before entering the prompt. They also agreed that it would be great if they could convert the output image into an editable format file in Adobe Illustrator or Photoshop.

\subsection{Limitations and Failure Cases}
 Many prompts are required by participants who want to change the perspective or transform a specific part of the object in the resulting image. This leads users to spend more prompts on iCONTRA to complete the tasks. Although we edit the image guided by masks, we still fail when editing the specific region of the object, requiring more complex details and unable to generate a different pose, view of the generated object or retain the original object. Additionally, users provide extremely detailed descriptions of the object, but the model does not completely understand them (as illustrated in Fig.~\ref{fig:failure_cases}).

Our method inherits most of the limitations of the pretrained Stable Diffusion in generating desired images and suffers from the following main aspects. We are unable to edit the specific region of objects, even small or non-salient ones. This problem arises because the model is trained on datasets like image-captioning, where the text prompt only attends to the salient object and generates only one object corresponding to the text prompt. Another issue arises from our heavy dependence on the image layout generated from the given prompt \(P_t\). If the Stable Diffusion model struggles to produce the desired layout or shape, our method encounters difficulties. Additionally, the dataset on which the model is trained lacks details of description, so the model is unable to understand prompt completely.


\section{Conclusion and Future Work}
We discussed with participants how to tackle these problems and get more insights for the next improvements of iCONTRA. For the need to modify a specific area in the picture, participants expressed a preference for the ability to select the region for alteration before entering the prompt. They also agreed that it would be great if they could convert the output image into an editable format file in Adobe Illustrator or Photoshop.

We introduced iCONTRA, a system enabling users to upload an input object image and generate related objects based on that input. Insights and feedback from the pilot study suggest that iCONTRA has partially demonstrated its capability to assist users in minimizing cognitive effort and formulating prompts for generating desired objects in the mentioned task.

In our forthcoming endeavors, our focus is on enhancing the visual guidance offered to users during image manipulation. This enhancement aims to provide users with greater control and personalization options, allowing them to select specific regions they want to modify or change the perspective they want to see. By enabling users to exert more influence over the image generation and editing process, we anticipate that the resulting outputs will be better aligned with their creative vision. To change the pose, view, edit specific regions and understand completely prompt, we must fine-tune the model on a new dataset that describes more complex details of objects. Ultimately, these improvements will contribute to a more tailored and satisfying experience for users.

\section*{Acknowledgments}

This research is funded by Vietnam National Foundation for Science and Technology Development (NAFOSTED) under Grant Number 102.05-2023.31. 

Additionally, Dr. Tam V. Nguyen is supported by National Science Foundation (NSF) under Grant Number 2025234.

\bibliographystyle{x/ACM-Reference-Format}
\bibliography{x/sample-base}


\begin{thebibliography}{20}


\ifx \showCODEN    \undefined \def \showCODEN     #1{\unskip}     \fi
\ifx \showDOI      \undefined \def \showDOI       #1{#1}\fi
\ifx \showISBNx    \undefined \def \showISBNx     #1{\unskip}     \fi
\ifx \showISBNxiii \undefined \def \showISBNxiii  #1{\unskip}     \fi
\ifx \showISSN     \undefined \def \showISSN      #1{\unskip}     \fi
\ifx \showLCCN     \undefined \def \showLCCN      #1{\unskip}     \fi
\ifx \shownote     \undefined \def \shownote      #1{#1}          \fi
\ifx \showarticletitle \undefined \def \showarticletitle #1{#1}   \fi
\ifx \showURL      \undefined \def \showURL       {\relax}        \fi
\providecommand\bibfield[2]{#2}
\providecommand\bibinfo[2]{#2}
\providecommand\natexlab[1]{#1}
\providecommand\showeprint[2][]{arXiv:#2}

\bibitem[Cao et~al\mbox{.}(2023)]%
        {cao_2023_masactrl}
\bibfield{author}{\bibinfo{person}{Mingdeng Cao}, \bibinfo{person}{Xintao Wang}, \bibinfo{person}{Zhongang Qi}, \bibinfo{person}{Ying Shan}, \bibinfo{person}{Xiaohu Qie}, {and} \bibinfo{person}{Yinqiang Zheng}.} \bibinfo{year}{2023}\natexlab{}.
\newblock \showarticletitle{MasaCtrl: Tuning-Free Mutual Self-Attention Control for Consistent Image Synthesis and Editing}. In \bibinfo{booktitle}{\emph{Proceedings of the IEEE/CVF International Conference on Computer Vision (ICCV)}}. \bibinfo{pages}{22560--22570}.
\newblock


\bibitem[Chefer et~al\mbox{.}(2023)]%
        {chefer2023attendandexcite}
\bibfield{author}{\bibinfo{person}{Hila Chefer}, \bibinfo{person}{Yuval Alaluf}, \bibinfo{person}{Yael Vinker}, \bibinfo{person}{Lior Wolf}, {and} \bibinfo{person}{Daniel Cohen-Or}.} \bibinfo{year}{2023}\natexlab{}.
\newblock \bibinfo{title}{Attend-and-Excite: Attention-Based Semantic Guidance for Text-to-Image Diffusion Models}.
\newblock
\newblock
\showeprint[arxiv]{2301.13826}~[cs.CV]


\bibitem[Elarabawy et~al\mbox{.}(2022)]%
        {elarabawy2022direct}
\bibfield{author}{\bibinfo{person}{Adham Elarabawy}, \bibinfo{person}{Harish Kamath}, {and} \bibinfo{person}{Samuel Denton}.} \bibinfo{year}{2022}\natexlab{}.
\newblock \bibinfo{title}{Direct Inversion: Optimization-Free Text-Driven Real Image Editing with Diffusion Models}.
\newblock
\newblock
\showeprint[arxiv]{2211.07825}~[cs.CV]


\bibitem[Hertz et~al\mbox{.}(2022)]%
        {hertz2022prompt}
\bibfield{author}{\bibinfo{person}{Amir Hertz}, \bibinfo{person}{Ron Mokady}, \bibinfo{person}{Jay Tenenbaum}, \bibinfo{person}{Kfir Aberman}, \bibinfo{person}{Yael Pritch}, {and} \bibinfo{person}{Daniel Cohen-Or}.} \bibinfo{year}{2022}\natexlab{}.
\newblock \showarticletitle{Prompt-to-Prompt Image Editing with Cross Attention Control}.
\newblock \bibinfo{journal}{\emph{arXiv preprint arXiv:2208.01626}} (\bibinfo{year}{2022}).
\newblock


\bibitem[Huang et~al\mbox{.}(2018)]%
        {huang2018multimodal}
\bibfield{author}{\bibinfo{person}{Xun Huang}, \bibinfo{person}{Ming-Yu Liu}, \bibinfo{person}{Serge Belongie}, {and} \bibinfo{person}{Jan Kautz}.} \bibinfo{year}{2018}\natexlab{}.
\newblock \showarticletitle{Multimodal unsupervised image-to-image translation}. In \bibinfo{booktitle}{\emph{Proceedings of the European conference on computer vision (ECCV)}}. \bibinfo{pages}{172--189}.
\newblock


\bibitem[Jing et~al\mbox{.}(2020)]%
        {8732370}
\bibfield{author}{\bibinfo{person}{Yongcheng Jing}, \bibinfo{person}{Yezhou Yang}, \bibinfo{person}{Zunlei Feng}, \bibinfo{person}{Jingwen Ye}, \bibinfo{person}{Yizhou Yu}, {and} \bibinfo{person}{Mingli Song}.} \bibinfo{year}{2020}\natexlab{}.
\newblock \showarticletitle{Neural Style Transfer: A Review}.
\newblock \bibinfo{journal}{\emph{IEEE Transactions on Visualization and Computer Graphics}} \bibinfo{volume}{26}, \bibinfo{number}{11} (\bibinfo{year}{2020}), \bibinfo{pages}{3365--3385}.
\newblock
\urldef\tempurl%
\url{https://doi.org/10.1109/TVCG.2019.2921336}
\showDOI{\tempurl}


\bibitem[Kawar et~al\mbox{.}(2023)]%
        {kawar2023imagic}
\bibfield{author}{\bibinfo{person}{Bahjat Kawar}, \bibinfo{person}{Shiran Zada}, \bibinfo{person}{Oran Lang}, \bibinfo{person}{Omer Tov}, \bibinfo{person}{Huiwen Chang}, \bibinfo{person}{Tali Dekel}, \bibinfo{person}{Inbar Mosseri}, {and} \bibinfo{person}{Michal Irani}.} \bibinfo{year}{2023}\natexlab{}.
\newblock \bibinfo{title}{Imagic: Text-Based Real Image Editing with Diffusion Models}.
\newblock
\newblock
\showeprint[arxiv]{2210.09276}~[cs.CV]


\bibitem[Kim et~al\mbox{.}(2022)]%
        {Kim_2022_CVPR}
\bibfield{author}{\bibinfo{person}{Gwanghyun Kim}, \bibinfo{person}{Taesung Kwon}, {and} \bibinfo{person}{Jong~Chul Ye}.} \bibinfo{year}{2022}\natexlab{}.
\newblock \showarticletitle{DiffusionCLIP: Text-Guided Diffusion Models for Robust Image Manipulation}. In \bibinfo{booktitle}{\emph{Proceedings of the IEEE/CVF Conference on Computer Vision and Pattern Recognition (CVPR)}}. \bibinfo{pages}{2426--2435}.
\newblock


\bibitem[Kirillov et~al\mbox{.}(2023)]%
        {kirillov2023segment}
\bibfield{author}{\bibinfo{person}{Alexander Kirillov}, \bibinfo{person}{Eric Mintun}, \bibinfo{person}{Nikhila Ravi}, \bibinfo{person}{Hanzi Mao}, \bibinfo{person}{Chloe Rolland}, \bibinfo{person}{Laura Gustafson}, \bibinfo{person}{Tete Xiao}, \bibinfo{person}{Spencer Whitehead}, \bibinfo{person}{Alexander~C. Berg}, \bibinfo{person}{Wan-Yen Lo}, \bibinfo{person}{Piotr Dollár}, {and} \bibinfo{person}{Ross Girshick}.} \bibinfo{year}{2023}\natexlab{}.
\newblock \bibinfo{title}{Segment Anything}.
\newblock
\newblock
\showeprint[arxiv]{2304.02643}~[cs.CV]


\bibitem[Le et~al\mbox{.}(2023)]%
        {le2023vides}
\bibfield{author}{\bibinfo{person}{Minh-Hien Le}, \bibinfo{person}{Chi-Bien Chu}, \bibinfo{person}{Khanh-Duy Le}, \bibinfo{person}{Tam~V. Nguyen}, \bibinfo{person}{Minh-Triet Tran}, {and} \bibinfo{person}{Trung-Nghia Le}.} \bibinfo{year}{2023}\natexlab{}.
\newblock \bibinfo{title}{VIDES: Virtual Interior Design via Natural Language and Visual Guidance}.
\newblock
\newblock
\showeprint[arxiv]{2308.13795}~[cs.CV]


\bibitem[Li et~al\mbox{.}(2017)]%
        {li2017demystifying}
\bibfield{author}{\bibinfo{person}{Yanghao Li}, \bibinfo{person}{Naiyan Wang}, \bibinfo{person}{Jiaying Liu}, {and} \bibinfo{person}{Xiaodi Hou}.} \bibinfo{year}{2017}\natexlab{}.
\newblock \bibinfo{title}{Demystifying Neural Style Transfer}.
\newblock
\newblock
\showeprint[arxiv]{1701.01036}~[cs.CV]


\bibitem[Liu et~al\mbox{.}(2017)]%
        {liu2017unsupervised}
\bibfield{author}{\bibinfo{person}{Ming-Yu Liu}, \bibinfo{person}{Thomas Breuel}, {and} \bibinfo{person}{Jan Kautz}.} \bibinfo{year}{2017}\natexlab{}.
\newblock \showarticletitle{Unsupervised image-to-image translation networks}.
\newblock \bibinfo{journal}{\emph{Advances in neural information processing systems}}  \bibinfo{volume}{30} (\bibinfo{year}{2017}).
\newblock


\bibitem[Mokady et~al\mbox{.}(2022)]%
        {mokady2022null}
\bibfield{author}{\bibinfo{person}{Ron Mokady}, \bibinfo{person}{Amir Hertz}, \bibinfo{person}{Kfir Aberman}, \bibinfo{person}{Yael Pritch}, {and} \bibinfo{person}{Daniel Cohen-Or}.} \bibinfo{year}{2022}\natexlab{}.
\newblock \showarticletitle{Null-text Inversion for Editing Real Images using Guided Diffusion Models}.
\newblock \bibinfo{journal}{\emph{arXiv preprint arXiv:2211.09794}} (\bibinfo{year}{2022}).
\newblock


\bibitem[Nichol et~al\mbox{.}(2022)]%
        {nichol2022glide}
\bibfield{author}{\bibinfo{person}{Alex Nichol}, \bibinfo{person}{Prafulla Dhariwal}, \bibinfo{person}{Aditya Ramesh}, \bibinfo{person}{Pranav Shyam}, \bibinfo{person}{Pamela Mishkin}, \bibinfo{person}{Bob McGrew}, \bibinfo{person}{Ilya Sutskever}, {and} \bibinfo{person}{Mark Chen}.} \bibinfo{year}{2022}\natexlab{}.
\newblock \bibinfo{title}{GLIDE: Towards Photorealistic Image Generation and Editing with Text-Guided Diffusion Models}.
\newblock
\newblock
\showeprint[arxiv]{2112.10741}~[cs.CV]


\bibitem[Parmar et~al\mbox{.}(2023)]%
        {parmar2023zeroshot}
\bibfield{author}{\bibinfo{person}{Gaurav Parmar}, \bibinfo{person}{Krishna~Kumar Singh}, \bibinfo{person}{Richard Zhang}, \bibinfo{person}{Yijun Li}, \bibinfo{person}{Jingwan Lu}, {and} \bibinfo{person}{Jun-Yan Zhu}.} \bibinfo{year}{2023}\natexlab{}.
\newblock \bibinfo{title}{Zero-shot Image-to-Image Translation}.
\newblock
\newblock
\showeprint[arxiv]{2302.03027}~[cs.CV]


\bibitem[Ramesh et~al\mbox{.}(2022)]%
        {ramesh2022hierarchical}
\bibfield{author}{\bibinfo{person}{Aditya Ramesh}, \bibinfo{person}{Prafulla Dhariwal}, \bibinfo{person}{Alex Nichol}, \bibinfo{person}{Casey Chu}, {and} \bibinfo{person}{Mark Chen}.} \bibinfo{year}{2022}\natexlab{}.
\newblock \bibinfo{title}{Hierarchical Text-Conditional Image Generation with CLIP Latents}.
\newblock
\newblock
\showeprint[arxiv]{2204.06125}~[cs.CV]


\bibitem[Rombach et~al\mbox{.}(2021)]%
        {rombach2021highresolution}
\bibfield{author}{\bibinfo{person}{Robin Rombach}, \bibinfo{person}{Andreas Blattmann}, \bibinfo{person}{Dominik Lorenz}, \bibinfo{person}{Patrick Esser}, {and} \bibinfo{person}{Björn Ommer}.} \bibinfo{year}{2021}\natexlab{}.
\newblock \bibinfo{title}{High-Resolution Image Synthesis with Latent Diffusion Models}.
\newblock
\newblock
\showeprint[arxiv]{2112.10752}~[cs.CV]


\bibitem[Rombach et~al\mbox{.}(2022)]%
        {rombach2022highresolution}
\bibfield{author}{\bibinfo{person}{Robin Rombach}, \bibinfo{person}{Andreas Blattmann}, \bibinfo{person}{Dominik Lorenz}, \bibinfo{person}{Patrick Esser}, {and} \bibinfo{person}{Björn Ommer}.} \bibinfo{year}{2022}\natexlab{}.
\newblock \bibinfo{title}{High-Resolution Image Synthesis with Latent Diffusion Models}.
\newblock
\newblock
\showeprint[arxiv]{2112.10752}~[cs.CV]


\bibitem[Ruiz et~al\mbox{.}(2023)]%
        {Ruiz_2023_CVPR}
\bibfield{author}{\bibinfo{person}{Nataniel Ruiz}, \bibinfo{person}{Yuanzhen Li}, \bibinfo{person}{Varun Jampani}, \bibinfo{person}{Yael Pritch}, \bibinfo{person}{Michael Rubinstein}, {and} \bibinfo{person}{Kfir Aberman}.} \bibinfo{year}{2023}\natexlab{}.
\newblock \showarticletitle{DreamBooth: Fine Tuning Text-to-Image Diffusion Models for Subject-Driven Generation}. In \bibinfo{booktitle}{\emph{Proceedings of the IEEE/CVF Conference on Computer Vision and Pattern Recognition (CVPR)}}. \bibinfo{pages}{22500--22510}.
\newblock


\bibitem[Saharia et~al\mbox{.}(2022)]%
        {saharia2022photorealistic}
\bibfield{author}{\bibinfo{person}{Chitwan Saharia}, \bibinfo{person}{William Chan}, \bibinfo{person}{Saurabh Saxena}, \bibinfo{person}{Lala Li}, \bibinfo{person}{Jay Whang}, \bibinfo{person}{Emily Denton}, \bibinfo{person}{Seyed Kamyar~Seyed Ghasemipour}, \bibinfo{person}{Burcu~Karagol Ayan}, \bibinfo{person}{S.~Sara Mahdavi}, \bibinfo{person}{Rapha~Gontijo Lopes}, \bibinfo{person}{Tim Salimans}, \bibinfo{person}{Jonathan Ho}, \bibinfo{person}{David~J Fleet}, {and} \bibinfo{person}{Mohammad Norouzi}.} \bibinfo{year}{2022}\natexlab{}.
\newblock \bibinfo{title}{Photorealistic Text-to-Image Diffusion Models with Deep Language Understanding}.
\newblock
\newblock
\showeprint[arxiv]{2205.11487}~[cs.CV]


\end{thebibliography}

\appendix

\end{document}